# Shapley variable importance clouds for interpretable machine learning

Yilin Ning[1], Marcus Eng Hock Ong[2,3,4], Bibhas Chakraborty[1,2,5,6], Benjamin Alan Goldstein[2,6], Daniel Shu Wei Ting[1], Roger Vaughan[1,2], Nan Liu[1,2,3,7,8*]

[1] Centre for Quantitative Medicine, Duke-NUS Medical School, Singapore, Singapore
[2] Programme in Health Services and Systems Research, Duke-NUS Medical School, Singapore, Singapore
[3] Health Services Research Centre, Singapore Health Services, Singapore, Singapore
[4] Department of Emergency Medicine, Singapore General Hospital, Singapore, Singapore
[5] Department of Statistics and Data Science, National University of Singapore, Singapore, Singapore
[6] Department of Biostatistics and Bioinformatics, Duke University, Durham, NC, United States
[7] SingHealth AI Health Program, Singapore Health Services, Singapore, Singapore
[8] Institute of Data Science, National University of Singapore, Singapore, Singapore

*Correspondence: Nan Liu, Programme in Health Services and Systems Research, Duke-NUS Medical School, 8 College Road, Singapore, 169857. Phone: +65 6601 6503. Email: liu.nan@duke-nus.edu.sg

## Abstract

Interpretable machine learning has been focusing on explaining final models that optimize performance. The current state-of-the-art is the Shapley additive explanations (SHAP) that locally explains variable impact on individual predictions, and it is recently extended for a global assessment across the dataset. Recently, Dong and Rudin proposed to extend the investigation to models from the same class as the final model that are "good enough", and identified a previous overclaim of variable importance based on a single model. However, this method does not directly integrate with existing Shapley-based interpretations. We close this gap by proposing a Shapley variable importance cloud that pools information across good models to avoid biased assessments in SHAP analyses of final models, and communicate the findings via novel visualizations. We demonstrate the additional insights gain compared to conventional explanations and Dong and Rudin's method using criminal justice and electronic medical records data.

## Main

Machine learning (ML) methods has been widely used to aid high-stakes decision making, e.g., in healthcare[1,2] and crime prediction.[3] While ML models achieve good performance by capturing data patterns through complex mathematical structures, such complexity results in "black box" models that hide the underlying mechanism. Inability to assess the connection between variables and predictions makes it difficult to detect potential flaws and bias in the resulting prediction models and limits their uptake in real-life decision making.[4–8] The growing research on interpretable machine learning (IML), also interchangeably referred to as explainable artificial intelligence in the literature, improves the usability of ML models by revealing the contribution of variables to predictions.[7–11]

A lot of effort in IML has been put into "post-hoc" explanations that quantifies variable impact on a model while leaving the model a black box.[8,10,11] For example, the random forest[12] was developed with a built-in permutation importance that evaluates reductions in model performance after removing each variable, which partially contributes to its wide adoption in practice.[11] Recently, this permutation approach was extended to a model-agnostic approach that provides global explanations for any ML models.[13] Current IML applications are dominated by two local model-agnostic explanation approaches:[14] the local interpretable model-agnostic explanations (LIME)[15] explains individual predictions by locally approximating them with interpretable models; the Shapley additive explanations (SHAP)[16] attributes a prediction among variables by considering it as a cooperative game, which avoids fitting surrogate models and accounts for interactions among variables.

A desirable property of SHAP is that in addition to locally explaining individual predictions, the mean absolute SHAP values can provide heuristic measures of variable importance to overall model performance,[16,17] and a formal global extension, i.e., Shapley additive global importance (SAGE),[17] is developed recently. However, by leaving the black box unopened these methods do not fully reveal the mechanism of the models, e.g., why do some variables contribute more to the predictions than others.[8] Ante-hoc IML methods address this by developing inherently interpretable models, e.g., recent works[18–20] proposed ML approaches to build sparse scoring systems based on simple regression models that had good discriminative ability. By integrating considerations such as variable importance into model building steps, these methods support direct inference on the importance of variables to the outcome.

While most IML approaches focus on optimal (e.g., loss-minimizing) models, a recent work[21] broadened the scope to a wider range of models that are "good enough". These nearly optimal models are highly relevant to practical questions, e.g., can an accurate yet expensive biomarker be replaced with other variable(s) without strongly impairing prediction accuracy?[21] To systematically address such questions, Dong and Rudin[21] proposed a variable importance cloud (VIC) that provides a comprehensive overview of variable contributions by analyzing the variability of variable importance across a group of nearly optimal models, and found an overclaim of the importance of race to criminal recidivism prediction in post-hoc assessments.

VIC demonstrated the benefit of extending global interpretation to nearly optimal models, but it was developed from the permutation importance,[13,21] hence leaving a gap between theoretical developments and current applications based on SHAP. We propose to extend the state-of-the-art SHAP method to higher-level global interpretations by integrating the latest development in Shapley-based variable importance measures with the recently proposed VIC framework. With our proposed approach, referred to as ShapleyVIC, we explicitly analyze the variability of variable importance across models, convey it through novel visualizations, and describe the practical implications as a complement to SHAP analysis. We validate ShapleyVIC by reproducing main results in the previous analysis of criminal recidivism prediction,[21] and demonstrate the use of the integrated SHAP-ShapleyVIC framework for a comprehensive assessment of variable contributions when predicting mortality using a real-life clinical data.

## Results

### Shapley variable importance cloud (ShapleyVIC)

Following the VIC framework, our proposed ShapleyVIC extends the widely used Shapley-based variable importance measures beyond final models for a comprehensive assessment and has important practical implications. We describe the permutation importance used in VIC and popular Shapley-based methods, define ShapleyVIC with explicit variability measures to support practical inference, and describe our practical solutions to some challenges in implementation.

**Global importance measures.** Let $Y$ denote the outcome and let $X_D = \{X_1, \dots, X_d\}$ collectively denote $d$ variables, where $D = \{1, \dots, d\}$ is the set of all variable indices. A model of $Y$ built using the $d$ variables is denoted by $f(X_D)$, with expected loss $E\{L(f(X_D), Y)\}$. Fisher and team[13] proposed a permutation-based measure of variable contribution, referred to as model reliance (MR). MR of variable $X_j$ ($j \in D$) is the increase in expected loss when the contribution of this variable is removed by random permutation:

$$mr_j(f) = \frac{E\{L(f(X_{D\setminus\{j\}}, X_j'), Y)\}}{E\{L(f(X_D), Y)\}},$$

where $X_{D\setminus\{j\}}$ denotes the set $X_D$ after excluding $X_j$, and $X_j'$ follows the marginal distribution of $X_j$. $mr_j(f) = 1$ suggests model $f$ does not rely on $X_j$, and larger $mr_j(f)$ indicates increased reliance.

Although straightforward and easy to implement, the permutation approach does not account for interactions among variables as it removes one variable at a time.[17,22] Shapley-based explanations account for this by viewing variables as players in a cooperative game,[16,17] and measures the impact of variable $X_j$ on model $f$ based on its marginal contribution when some variables, $X_S \subset X_D$, are already present:

$$\phi_j(w) = \frac{1}{d} \sum_{S \subseteq \{D \setminus \{j\}\}} \binom{d-1}{|S|}^{-1} [w(S \cup \{j\}) - w(S)]. \quad (1)$$

$w(S)$ denotes the contribution of subset $X_S$, $|S|$ denotes the number of variables in this subset, and $\binom{d-1}{|S|}$ is the number of ways to choose $|S|$ variables from $X_{D\setminus\{j\}}$. $\phi_j(w) = 0$ indicates no contribution, and larger values indicate increased contribution.[17]

When $w(S)$ is the expectation of a single prediction, i.e., $w(S) = v_{f,x}(S) = E[f(X_D|X_S = x_S)]$, $\phi_j(v_{f,x})$ gives the SHAP value for local explanation.[16] Absolute SHAP values reflect the magnitude of variable impact and the signs indicate the direction, therefore the mean absolute SHAP value may be used as a heuristic global importance measure.[16,17]

When $w(S)$ is the expected reduction in loss over mean prediction by including $X_S$, i.e., $w(S) = v_f(S) = E\{L(E[f(X_D)], Y)\} - E\{L(f(X_D|X_S = x_S), Y)\}$, $\phi_j(v_f)$ is the SAGE value for a formal global interpretation.[17] Our proposed ShapleyVIC follows the VIC approach to extend the global and model-agnostic SAGE across models.

**ShapleyVIC definition.** Suppose $f^*(X_D)$ is the optimal model that minimizes expected loss among all possible $f$. Dong and Rudin[21] proposed to extend the investigation of variable importance to a Rashomon set of models with nearly-optimal performance (in terms of expected loss):

$$R(\varepsilon, f^*, F) = \left\{f \in F \middle| E\{L(f(X_D), Y)\} \le (1+\varepsilon) E\{L(f^*(X_D), Y)\}\right\},$$

where "nearly-optimal" is defined by the small positive value $\varepsilon$, e.g., $\varepsilon = 5\%$. Using $MR(f) = \{mr_1(f), \dots, mr_d(f)\}$ to denote the collection of MR for model $f$, the variable importance cloud (VIC) is the collection of model reliance functions of all models in the Rashomon set defined above:[13,21]

$$VIC(R) = \{MR(f) : f(X_D) \in R(\varepsilon, f^*, F)\}.$$

VIC values are asymptotically normally distributed, but calculating their standard error (SE) is non-trivial when $f$ is not a linear regression model.[13,21]

Our proposed ShapleyVIC is a hybrid of ante-hoc and post-hoc approaches, where the MR for each model in the Rashomon set is based on SAGE values. When working with models that are sensitive to collinearity among variables (e.g., regression models), we hypothesize that negative SAGE values with large absolute values are artifacts induced by highly correlated variables rather than indications of unimportance. Therefore, we define the Shapley-based MR based on the variance inflation factor[23] (VIF) of each variable:

$$mr_j^s(f) = \begin{cases} \left|\phi_j(v_f)\right| \text{ if } VIF_j \geq 2, \\ \phi_j(v_f) \text{ if } VIF_j < 2, \end{cases}$$

where $j = 1, \ldots, d$, and the superscript $s$ indicates the Shapley-based approach. Colinear variables will have similar MR values. The corresponding ShapleyVIC is:

$$VIC^S(R) = \{MR^S(f) : f(X_D) \in R(\varepsilon, f^*, F)\},$$

where $MR^S(f) = \{mr_1^s(f), \ldots, mr_d^s(f)\}$.

**ShapleyVIC inference.** Only positive ShapleyVIC values indicate importance, and larger value suggests higher importance. A desirable property of ShapleyVIC is that the SE of each value is readily available from the SAGE algorithm: $\sigma_j(f) = SE\left(\widehat{mr}_j^s(f)\right) = SE\left(\hat{\phi}_j(v_f)\right)$. This allows us to easily compare the reliance of a model on any two variables, $\{X_j, X_k\} \in X_D$, where the difference is normally distributed with variance $var\{\widehat{mr}_j^s(f) - \widehat{mr}_k^s(f)\} = \sigma_j^2(f) + \sigma_k^2(f)$ (assuming independence between $mr_j^s(f)$ and $mr_k^s(f)$). The importance of the $d$ variables to the model can be ranked based on the number of times each variable has significantly larger ShapleyVIC value than the other $d - 1$ variables.

To assess the overall importance of $X_j$ over the Rashomon set, we adopt a meta-analysis approach by viewing each model as a separate study. Since each model has a different true reliance on $X_j$, we use the random effect approach,[24–26] assuming that these true values are normally distributed around a grand mean. The variability of ShapleyVIC values is decomposed into a within-model variance (i.e., $\sigma_j^2(f)$) and a between-model variance (i.e., the variance of true values, to be estimated using standard meta-analysis approach[24,26]). The inverse total variances are used as weights to compute the average ShapleyVIC value, which estimates the overall importance of variables, and the 95% prediction interval[25] (PI) for a new model from the Rashomon set is used to statistically assess and compare overall importance. Mathematical details are provided in the Methods section.

**ShapleyVIC implementation.** Although VIC is validly computed from the same data used to train the optimal model,[21] we adopt the approach in SHAP and SAGE[17,27] to evaluate ShapleyVIC values using the testing set, and use the training set to train the optimal model and identify the Rashomon set. Larger sample requires longer computation time,[17,28] therefore we do not recommend using testing sets larger than necessary for the algorithm to converge.

As a hybrid of model-agnostic VIC and SAGE, ShapleyVIC is also model agnostic. In view of the popularity of scoring models, which are often built upon logistic regression models for binary outcomes, we describe the implementation of ShapleyVIC with logistic regression models. In such scenarios, the Rashomon set consists of regression coefficients, $\boldsymbol{\beta}$, corresponding to expected logistic loss $E\{L\} \leq (1+\varepsilon)E\{L^*\}$, where the superscript "*" indicates the optimal model with minimum expected loss, and $\varepsilon = 5\%$ is an acceptable value. Dong and Rudin drew samples of $\boldsymbol{\beta}$ via an ellipsoid approximation to the Rashomon set, which worked well in their examples but may not generalize to higher dimensions and/or stronger correlations.[21] We consider a pragmatic sampling approach based on rejection sampling:

- Set initial values for $M_0$ (the number of initial samples to draw from the Rashomon set), and $u_1$ and $u_2$ (bounds of a uniform distribution).
- For each $i = 1, \ldots, M_0$, generate $k_i \sim U(u_1, u_2)$.
- Draw the $i$-th sample from a multivariate normal distribution: $\boldsymbol{\beta}_i \sim N(\boldsymbol{\beta}^*, k_i \Sigma^*)$, where $\boldsymbol{\beta}^*$ is the regression coefficients of the optimal model and $\Sigma^*$ is its variance-covariance matrix. Reject $\boldsymbol{\beta}_i$ if the corresponding empirical loss, $\hat{L}_i$, exceeds the upper bound, i.e., if $\hat{L}_i > (1 + \varepsilon)\hat{L}^*$.
- Adjust the values of $M_0$, $u_1$ and $u_2$ such that the range between $\hat{L}^*$ and $(1 + \varepsilon)\hat{L}^*$ is well represented.

Following the practice of Dong and Rudin, we randomly select a final sample of 300 to 400 models.

**Experiments**

We used two data examples to demonstrate the implementation of ShapleyVIC and describe our proposed visualizations. In the first experiment, we validate ShapleyVIC by reproducing key findings in the COMPAS study of Dong and Rudin.[21,29] Next, we illustrate the use of ShapleyVIC as a complement to the widely used SHAP in an application for electronic health records.

**Experiment 1: COMPAS study.** This study aimed to assess the importance of six binary variables for predicting 2-year recidivation: age (dichotomized at 20 years), race (African American or others), having prior criminal history, having juvenile criminal history, and current charge (degree misdemeanor or others), with particular interest in race. We randomly divided the 7214 records into a training set with 90% (n=6393) records and a test set with the other 10% (n=721) records, and generated 350 nearly optimal logistic regression models.

This study illustrates the application of ShapleyVIC to a low-dimensional data with mild correlation (VIF<1.1 for all six variables based on the optimal model; see Figure 1(A)). We first assessed overall variable importance by inspecting the bar plot of the average ShapleyVIC values (with 95% PI) across the 350 models (see Figure 1(B)). Juvenile and prior criminal history had similar overall importance that were significantly higher than the other four variables, among which only age had significant overall importance (i.e., with 95% PI entirely above zero). The 95% PI of race contained zero, suggesting the same finding as Dong and Rudin[21] that generally race is not an important predictor of 2-year recidivation.

Inference on the bar plot of overall importance alone may lead to a misperception that variable importance is somewhat static. We qualitatively convey the variability across models by visualizing the relationship between model reliance on each variable and model performance using a colored violin plot (see Figure 1(C)). The horizontal spread of a violin represents the range of model reliance on a variable, which is divided into slices of equal width. The height of each slice represents the proportion of models in the model reliance interval, and the color indicates the average performance (in terms of empirical loss) of these models. If a model reliance interval does not contain any model (e.g., near the ends), the corresponding slice is combined with the neighbor closer to the center.

A key message from the violin plot is that there is no simple relationship between model performance and reliance on any variable, regardless of its overall variable importance, as illustrated by the well-mixed color across the range (see Figure 1(C)). For race, most dark-colored strips in the violin plot are positioned at negative model reliance values, suggesting its low contribution to better performing models. Bar and violin plots of VIC values from the same 350 models (see Appendix Figure A1) suggested similar findings but without variability measures to statistically test and compare overall importance.

**Experiment 2: MIMIC study.** We examined the importance of 21 variables (including age, clinical tests, and vital signs; see Figure 2(A) for a full list) in predicting 24-hour mortality in intensive care units (ICU), using a random sample of 20,000 adult patients from the BIDMC data set of the Medical Information Mart for Intensive Care (MIMIC) III database. We trained a logistic regression and generated a sample of 350 nearly optimal models using a random sample of 17,000 records, and used the rest 3000 records to evaluate variable importance. We used the SHAP-ShapleyVIC framework to assess the contribution of the 21 variables, beginning with a conventional SHAP analysis of the optimal logistic regression and followed by a ShapleyVIC assessment of nearly optimal models for additional insights.

*SHAP analysis of optimal model.* The extremely small p-values (<0.001; see Figure 2(A)) for two-thirds of the variables and collinearity among variables (indicated by large VIF values in Figure 2(A) and strong correlations in Appendix Figure A2) made it difficult to rank variable importance based on the optimal model. SHAP analysis of the model enabled straightforward variable ranking using mean absolute SHAP values (see Figure 2(B)). Per-instance SHAP values (indicated by dots in Figure 2(C)) provided additional insights on variable contribution to the optimal model, e.g., although creatinine only ranked 14$^{th}$ among the 21 variables, high creatinine levels can have a strong impact on predictions. Unlike SAGE, variances of SHAP values are not easily available for statistical assessments.

*ShapleyVIC analysis of nearly optimal models.* SHAP analysis of the optimal model does not answer some practical question, e.g., is creatinine deemed to contribute moderately to general prediction of ICU mortality using logistic regression? This is answered by the extended global interpretation using ShapleyVIC.

Figure 2(D) presents variable ranking after accounting for the variability in variable importance across the 350 nearly optimal models. Six of the top seven variables based on mean absolute SHAP values were also ranked top seven by average ShapleyVIC values, but ShapleyVIC tended to rank rest of the variables differently. The 95% PIs of average ShapleyVIC values suggested similar importance for the $5^{th}$- to $7^{th}$-ranking variables and statistically non-significant overall importance for the last five variables. VIC analysis of the 350 models had similar findings as ShapleyVIC on top-ranking variables (see Appendix Figure A3). The non-trivial relationship between variable contribution and model performance (indicated by the mixed color of the violin plot in Figure 2(E)) helps explore alternative well-performing models with higher reliance on creatinine.

Generally, creatinine contributed significantly to nearly optimal models, and ranked $13^{th}$ based on the average ShapleyVIC value (see Figure 2(D)). However, the violin plot (see Figure 2(E)) showed a wide spread of ShapleyVIC values for creatinine across models, and the dark blue strip at the right end suggested the presence of well-performing models that relied heavily on creatinine. To extract such models, we inspected variable ranking in each of the 350 models by pairwise comparison of ShapleyVIC values (visually summarized in Figure 3), and found creatinine ranked top seven in 19 models. Among these 19 models, creatinine increased to the $6^{th}$-ranking variable (see Figure 4), hemoglobin and hematocrit had lower ranks, while other variables were not much affected. Additional studies on creatinine may draw additional samples from the Rashomon set that are close to these 19 models for closer investigation.

## Discussion

Uncertainty is drawing attention when interpreting machine learning models,[11,30] which is relevant when interpretating predictions or estimated effects (e.g., see[31–33]) and when assessing the importance of variables (e.g., see[34,35]). We contribute to the investigation of uncertainty from a largely neglected source: the uncertainty in variable importance among nearly optimal models (e.g., where model loss is within an acceptable range) that could have been selected in a prediction task for practical considerations. By actively investigating the association between model performance and reliance on each variable, we provide a higher-level global assessment that studies model ensembles to avoid bias towards a single model when inferring variable importance (or unimportance), and provide a basis for building interpretable models under practical considerations and constraints.

The recently proposed VIC[21] is the first to demonstrate the benefit of extending global variable importance assessment to nearly optimal models. Our proposed method, named ShapleyVIC, is a hybrid of state-of-the-art ante-hoc and post-hoc IML approaches that extends the widely used Shapley-based explanations to global interpretations beyond a single optimal model. Using the meta-analysis approach, we pool the Shapley-based importance (measured by SAGE with 95% CI) from each model to explicitly quantify the uncertainty across models and summarize the overall importance of each variable. This allows us to support inference on variable importance with statistical evidence, which is not easily available from VIC.[21] The close connection between SHAP and SAGE[17,22] enables a seamless integration of ShapleyVIC with the state-of-the-art SHAP method for additional insight on variable contributions. Our proposed visualizations effectively communicates different levels of information and work well for high-dimensional data. We implemented ShapleyVIC as an R package, available from https://github.com/nliulab/ShapleyVIC.

Our experiment on the COMPAS data provides a strong motivation for extending global interpretation beyond a single model, where ShapleyVIC found that the importance of race in predicting recidivism in a post-hoc assessment was likely a random noise. The MIMIC study demonstrates the application of our proposed SHAP-ShapleyVIC framework. SHAP analysis of the optimal model facilitates straightforward interpretation of variable contribution, and subsequent ShapleyVIC analysis of nearly optimal models update the assessment by accounting for the variability in variable importance. By identifying variables with similar overall importance based on the variability between models, ShapleyVIC adds flexibility to model building steps, e.g., by considering stepwise inclusion or exclusion of such variables. Using our proposed visualizations of ShapleyVIC values across models, we demonstrated how to identify the presence of models with higher reliance on a variable of interest and subsequently focus on the relevant subset of models for additional information.

In common with VIC, ShapleyVIC faces a challenge in drawing representative samples of nearly optimal models due to the difficulty in characterizing the Rashomon set.[13,21,30] The more disciplined ellipsoid approximation approach described by Dong and Rudin may not work for high-dimensional and/or highly correlated data.[21] Our pragmatic sampling approach is applicable in such scenarios, but it may not preserve the asymptotic properties based on the Rashomon set.[13,21] However, by using the standard deviation of the Shapley-based model reliance, we are able to pool information across sampled models even when such asymptotic properties do not hold.

Strong correlation among variables (e.g., in the MIMIC study) also poses a challenge on variable importance assessment. Permutation importance is susceptible to biases when applied to correlated data as it samples from the marginal distribution.[17] SAGE is defined using the conditional distribution to account for correlations, but due to the immense computational challenge the authors adopted a sampling-based approximation approach that generates variables from marginal distributions and consequently assumes some extent of independence.[17] Similar challenges are encountered by other practical implementations of Shapley-based methods (e.g., see[22,36]) and are not easily resolved. By using the absolute value of SAGE as a measure of model reliance for variables with VIF>2, we provide a pragmatic solution to this problem that may inspire a more disciplined solution. ShapleyVIC may also be used with other (global) variable importance measures for preferrable properties.

In conclusion, in this study we present ShapleyVIC, a hybrid of the state-of-the-art ante-hoc and post-hoc IML approaches, which comprehensively assesses variable importance by extending the investigation to nearly optimal models that are relevant to practical prediction tasks. ShapleyVIC seamlessly integrates with SHAP due to the common theoretical basis, extending current IML applications to global interpretation and beyond. Although we described the implementation of ShapleyVIC with simple regression models, which can be readily integrated with development of scoring models (e.g., the recently developed AutoScore framework[20]), ShapleyVIC is model-agnostic and applicable for other machine learning models.

## Methods

### Pooling ShapleyVIC values using random effects meta-analysis

In this section, we describe our proposed method of pooling the Shapley-based model reliance of the $j$-th variable across $M$ nearly optimal models. The ShapleyVIC value of this variable for the $m$-th model and its variance are available from SAGE, denoted by $mr_{jm}^{\mathrm{s}}$ and $\sigma_{jm}^2$, respectively. To simplify notation, in this section we drop the subscript $j$.

Let $\theta_m$ denote the true ShapleyVIC value for the $m$-th model, where $mr_m^{\mathrm{s}} \sim N(\theta_m, \sigma_m^2)$. Since different models have different coefficients for variables and therefore different levels of reliance on each variable, $\theta_m$ is expected to differ across models. Hence, we adopt the random effects approach in meta-analysis[24–26] and assume a normal distribution for the true model reliance, $\theta_m \sim N(\theta, \tau^2)$, where the grand mean across models, $\theta$, and the between-model variability, $\tau^2$, are to be estimated.

We estimate $\tau^2$ using the commonly used DerSimonian-Laird approach.[24,26] The between-model variability ($\tau^2$) and within-model variability ($\sigma_m^2$, $m = 1, \dots, M$) are two sources of the total variance ($Q$), which is the weighted average of the squared deviation of $mr_m^{\mathrm{s}}$ from its weighted average: $Q = \sum w_m \{mr_m^{\mathrm{s}} - (\sum w_m mr_m^{\mathrm{s}})/(\sum w_m)\}^2$, and the weight is the inverse within-model variance: $w_m = 1/\sigma_m^2$. When within-model variability is the only source of total variance, $Q$ is expected to be $M - 1$. Hence, when $Q > M - 1$ the between-model variance can be estimated by $\tau^2 = (Q - (M - 1))/C$, where $C = (\sum w_m) - (\sum w_m^2)/(\sum w_m)$ is a scaling constant. If $Q \leq M - 1$ the estimated between-model variance is simply $\tau^2 = 0$.

With the estimated between-model variance, $\tau^2$, the grand mean, $\theta$, is estimated by a weighted average of $mr_m^s$: $\overline{mr}^s = (\sum w_m' mr_m^s)/(\sum w_m')$ and $var(\overline{mr}^s) = 1/(\sum w_m')$, where the updated weight is now the inverse of total variance, i.e., $w_m' = 1/(\sigma_m^2 + \tau^2)$.[24] To anticipate the ShapleyVIC value from a new model within the Rashomon set, $mr_{new}^s$, Higgins and team[25] proposed to assume a t-distribution with $M - 2$ degrees of freedom for $(mr_{new}^s - \overline{mr}^s)/\sqrt{(var(\overline{mr}^s) + \tau^2)}$. The 95% prediction interval for $mr_{new}^s$ is hence the 2.5-th and 97.5-th percentiles of this t-distribution. The estimation of the combined effect and the 95% prediction interval described in this section are implemented by the *metagen* function of the R package *meta*,[37] by using the default DerSimonian-Laird estimator for $\tau^2$ (*method="DL"*) and specifying *prediction=TRUE*.

**Figure 1.** Visual summary of COMPAS study results from the ShapleyVIC analysis. (A) The optimal logistic regression model, where variance inflation factors (VIF) did not suggest strong correlation among variables. (B) Overall importance of variables based on the average ShapleyVIC values from the 350 nearly optimal models. Variables with the 95% prediction interval containing zero or entirely below zero have non-significant overall importance (indicated by grey color). (C) Distribution of variable importance (indicated by the shape of violin plots) and the corresponding model performance.

**(A)** Optimal logistic regression model with minimum loss.

| Variables | Estimated coefficients | Standard error | P-value | VIF |
|---|---|---|---|---|
| Intercept | 0.445 | 0.107 | <0.001 | -- |
| Prior conviction | -0.854 | 0.061 | <0.001 | 1.06 |
| History of juvenile crime | -0.865 | 0.084 | <0.001 | 1.04 |
| Age | 1.500 | 0.187 | <0.001 | 1.03 |
| Race | 0.416 | 0.053 | <0.001 | 1.02 |
| Current charge | -0.254 | 0.056 | <0.001 | 1.02 |
| Gender | 0.384 | 0.068 | <0.001 | 1.01 |

**(B)** Variable ranking based on average ShapleyVIC values from 350 nearly optimal models.

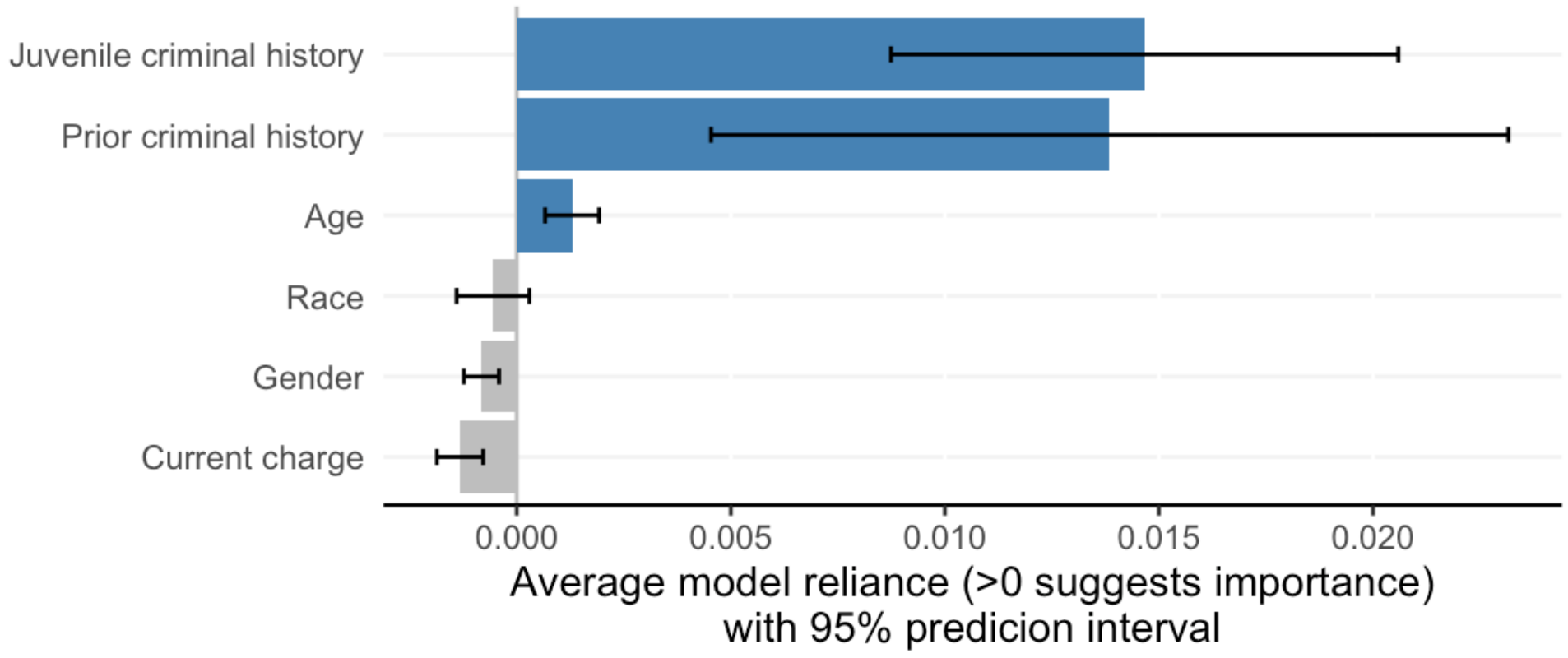

**(C)** ShapleyVIC values from 350 nearly optimal models.

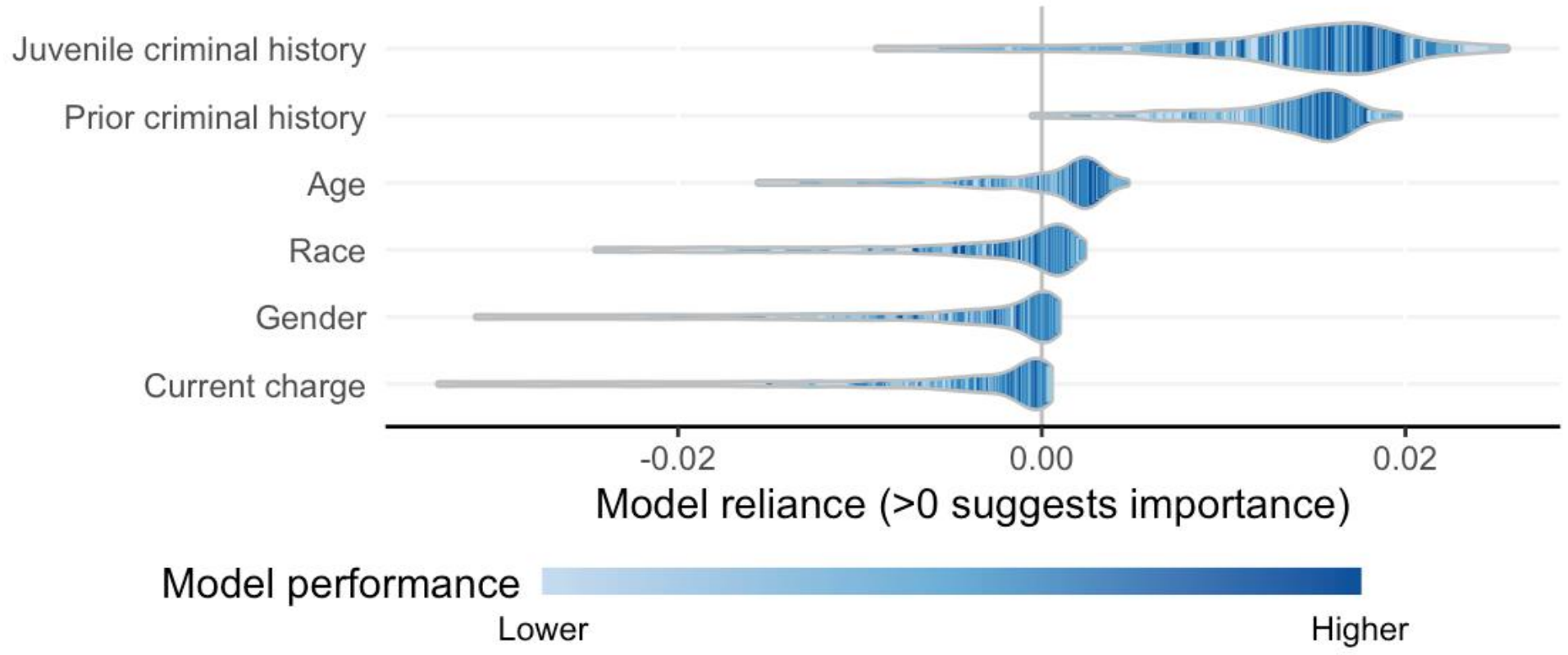

**Figure 2.** Visual summary of MIMIC study results from SHAP-ShapleyVIC framework. (A) The optimal logistic regression model, where high variance inflation factors (VIF>2) suggested strong correlation (indicated by grey color). (B) Variable ranking based on mean absolute SHAP values from the optimal model. (C) SHAP values (represented by dots) indicate variable contributions to individual predictions. (D) ShapleyVIC suggested a different variable ranking after accounting for the variability in variable importance across the 350 nearly optimal models. (E) Distribution of variable importance (indicated by the shape of violin plots) and the corresponding model performance (indicated by color) to supplement inference on average ShapleyVIC values. Dark blue strip towards the right end of a violin plot suggests the presence of good models that relied heavily on the variable (e.g., creatinine) for further investigations.

**(A)** Optimal logistic regression model with minimum expected loss.

| Variable | Estimated coefficient | Standard error | P-value | VIF |
|---|---|---|---|---|
| Intercept | 1.675 | 2.120 | 0.430 | -- |
| Hematocrit (%) | 0.078 | 0.020 | <0.001 | 15.57 |
| Hemoglobin (g/dL) | -0.296 | 0.057 | <0.001 | 15.57 |
| Chloride (mEq/L) | -0.037 | 0.015 | 0.013 | 12.92 |
| Bicarbonate (mmol/L) | -0.028 | 0.016 | 0.078 | 8.78 |
| Sodium (mmol/L) | 0.018 | 0.015 | 0.246 | 8.29 |
| Mean arterial pressure (MAP; mm Hg) | 0.029 | 0.007 | <0.001 | 7.92 |
| Anion gap (mEq/L) | 0.057 | 0.017 | 0.001 | 6.64 |
| Diastolic blood pressure (DBP; mm Hg) | -0.024 | 0.006 | <0.001 | 4.82 |
| Systolic blood pressure (SBP; mm Hg) | -0.016 | 0.003 | <0.001 | 3.17 |
| Creatinine (μmol/L) | -0.128 | 0.025 | <0.001 | 2.16 |
| Blood urea nitrogen (BUN; mg/dL) | 0.017 | 0.001 | <0.001 | 2.07 |
| Heart rate (beats/min) | 0.016 | 0.002 | <0.001 | 1.49 |
| Lactate (mmol/L) | 0.249 | 0.019 | <0.001 | 1.33 |
| Potassium (mmol/L) | -0.171 | 0.048 | <0.001 | 1.31 |
| Age (years) | 0.026 | 0.002 | <0.001 | 1.27 |
| Respiration (breaths/min) | 0.071 | 0.007 | <0.001 | 1.24 |
| Temperature (°C) | -0.077 | 0.047 | 0.098 | 1.20 |
| Platelet (thousand per microliter) | -0.001 | 0.0003 | <0.001 | 1.16 |
| Peripheral capillary oxygen saturation ($SpO_2$; %) | -0.030 | 0.012 | 0.010 | 1.12 |
| White blood cells (WBC; thousand per microliter) | 0.014 | 0.003 | <0.001 | 1.08 |
| Glucose (mg/dL) | 0.001 | 0.001 | 0.208 | 1.07 |

**SHAP analysis of the optimal model**

**(B)** Variable ranking based on optimal model.

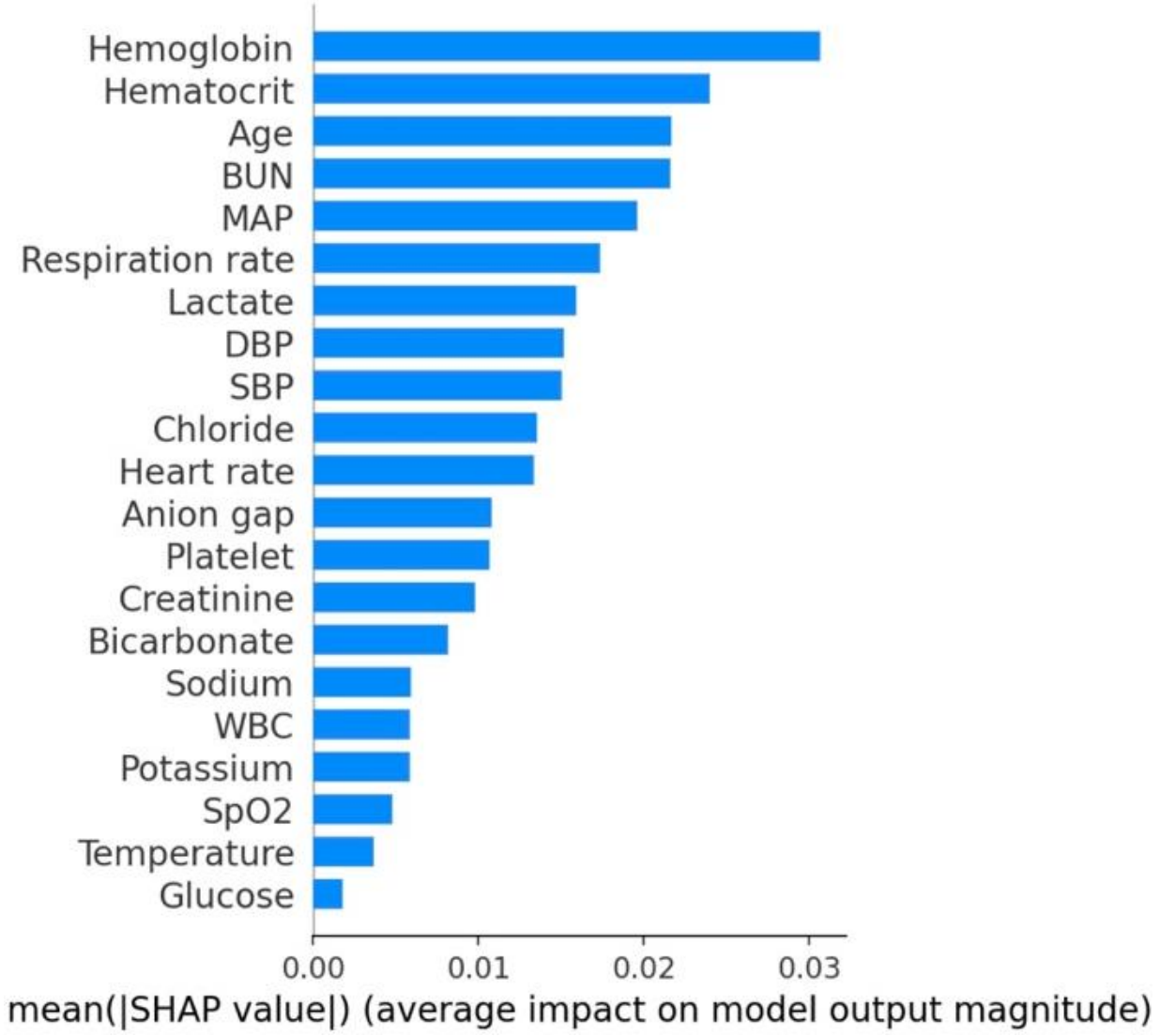

**(C)** Variable contributions to optimal model.

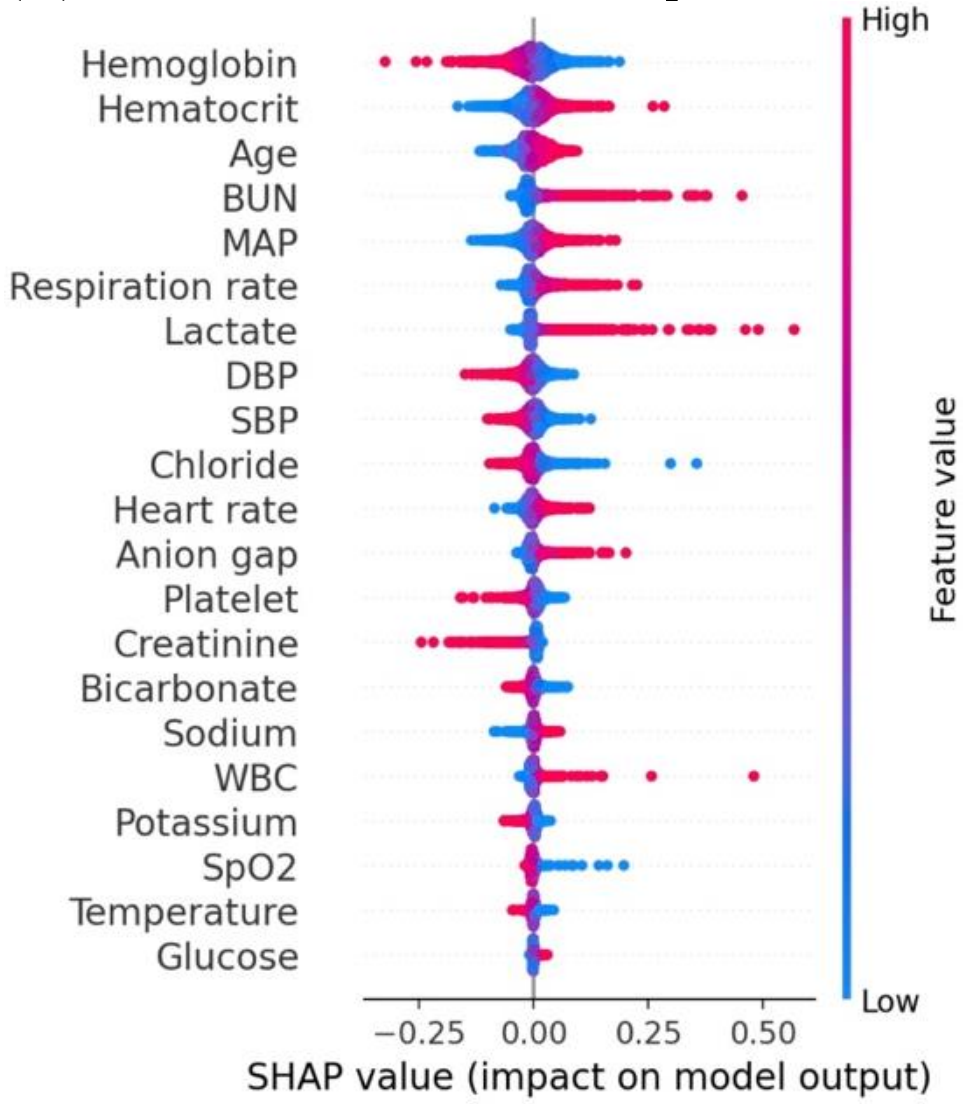

**ShapleyVIC analysis of 350 nearly optimal models**

**(D)** Variable ranking based on 350 nearly optimal models.

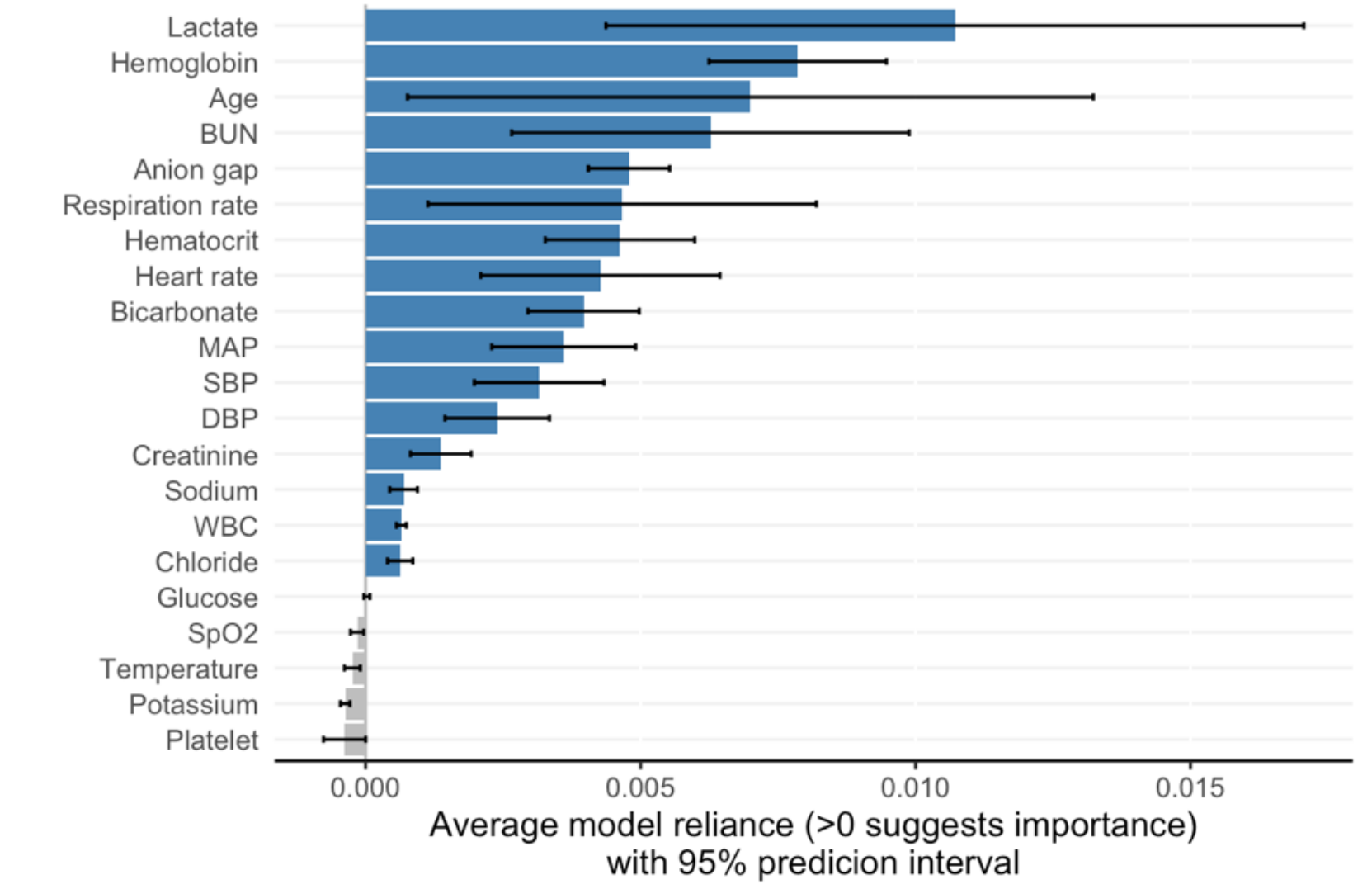

**(E)** Variable importance to 350 nearly optimal models.

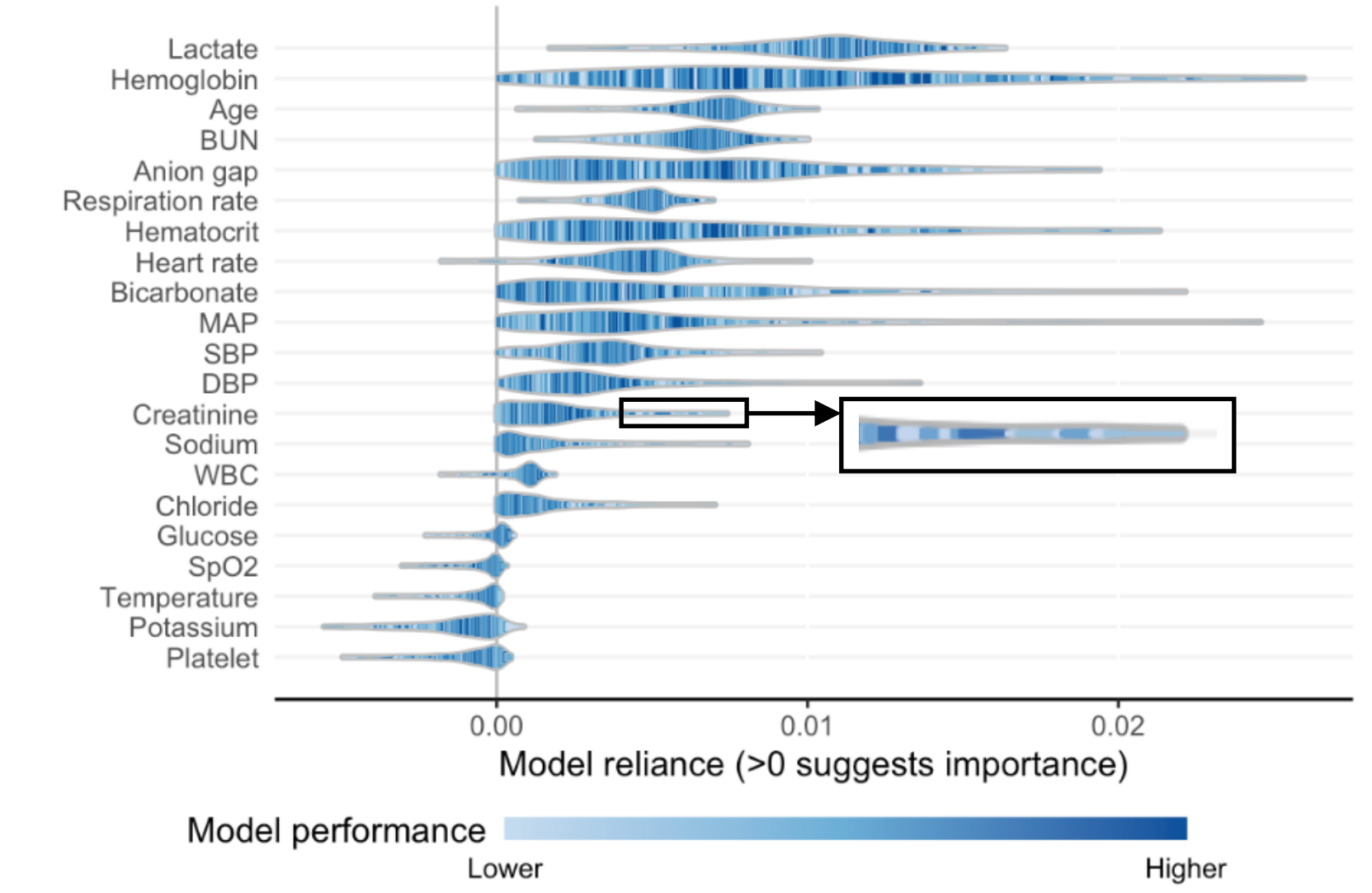

Figure 3. Frequency of ranking of each variable in the MIMIC study based on pairwise comparison of model reliance. Variables are arranged by average ShapleyVIC values.

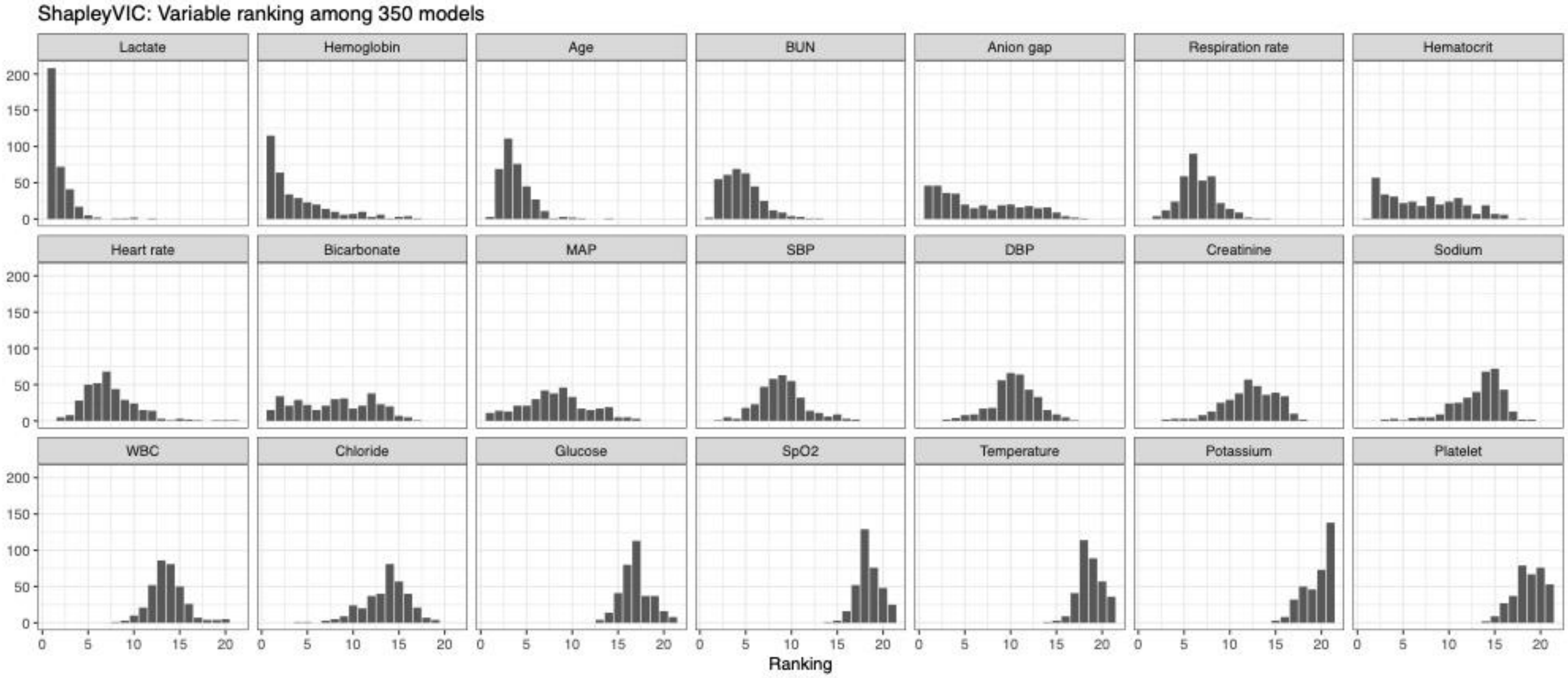

Figure 4. Bar plot of average ShapleyVIC values from 19 models where creatinine ranked top 7.

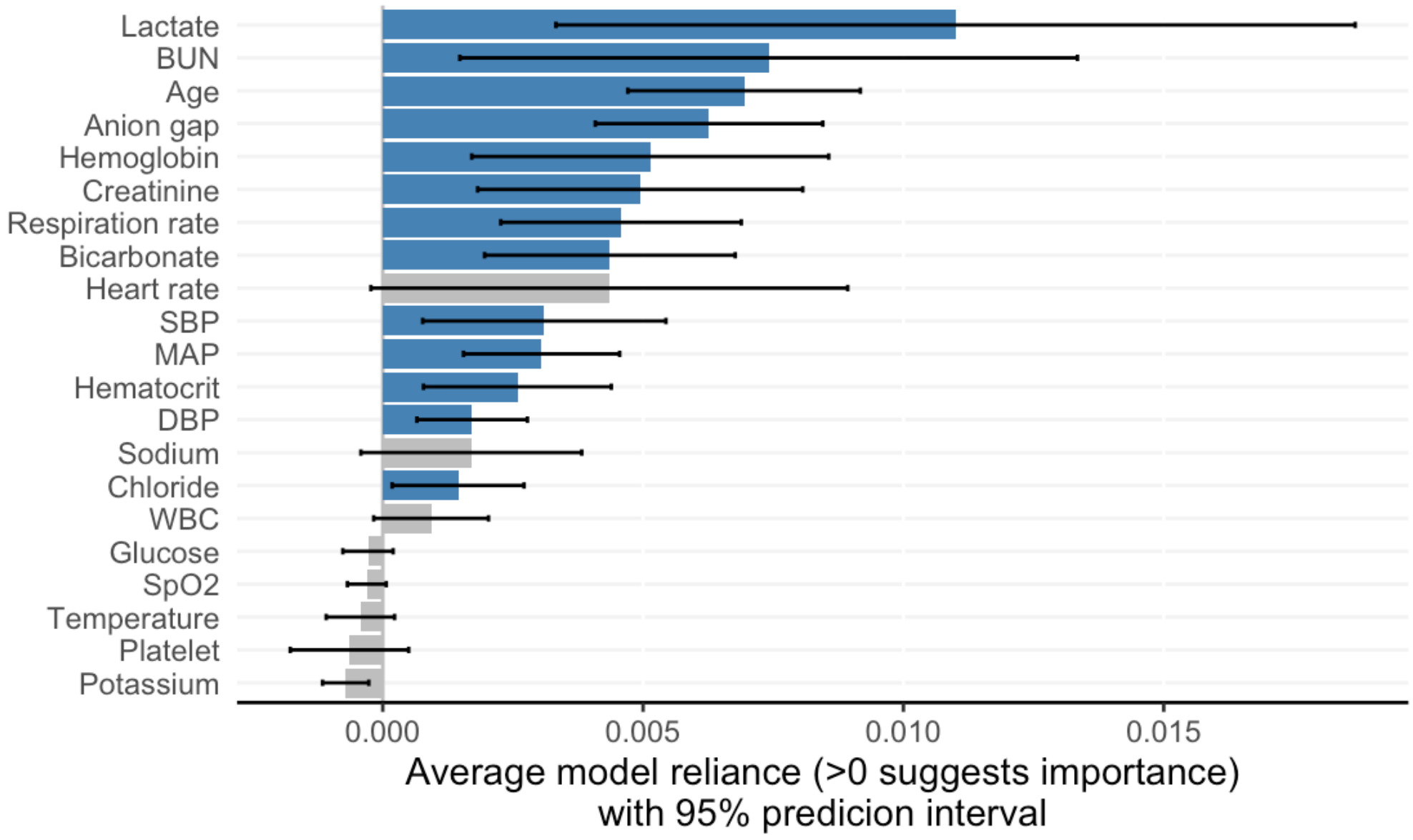

**Appendix**

**Figure A1.** Visual summary of COMPAS study results from the VIC analysis. VIC values were subtracted 1 such that zero value or lower indicate unimportance. (A) Average values from the 350 nearly optimal models. Variable ranking based on average VIC values is similar to that based on ShapleyVIC values. (B) Distribution of VIC values (subtracted 1) and the corresponding model performance.

**(A)** Variable ranking based on average VIC values (subtracted 1) from 350 nearly optimal models.

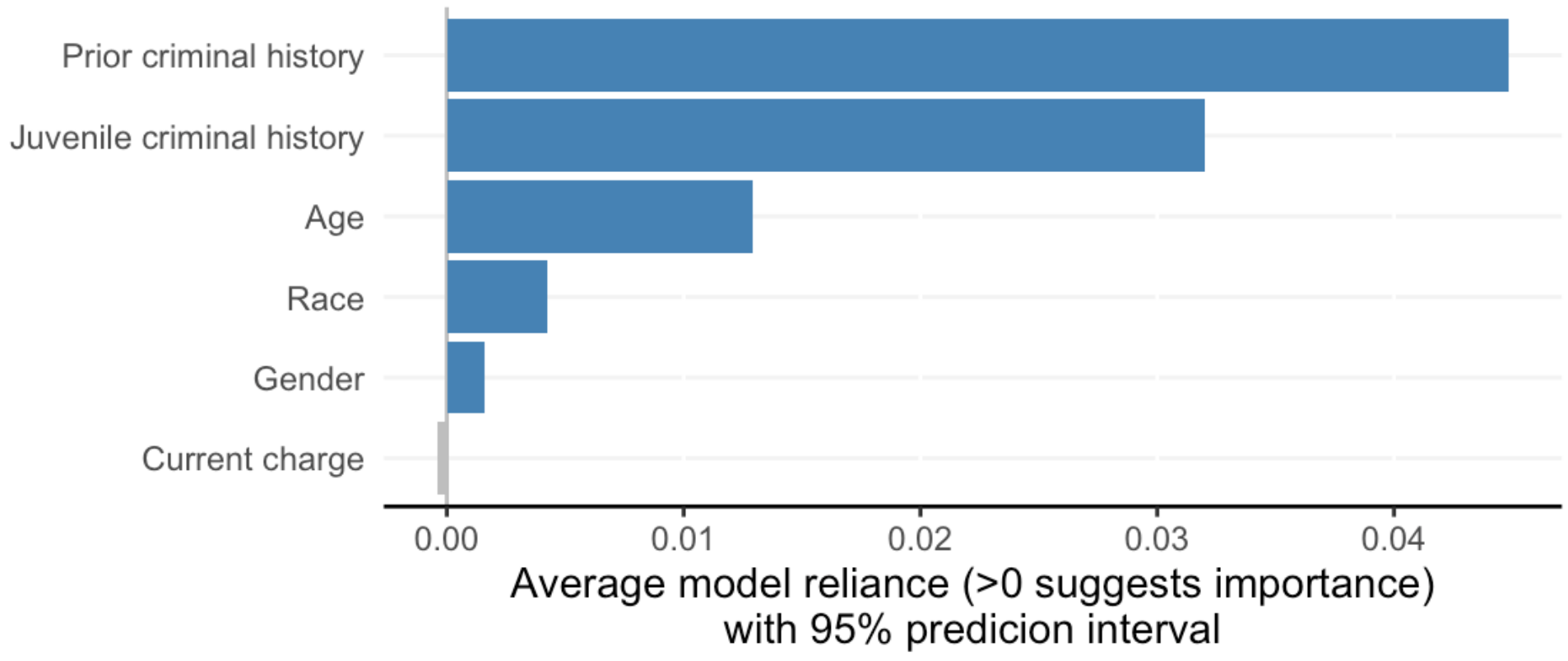

**(B)** VIC values (subtracted 1) from 350 nearly optimal models.

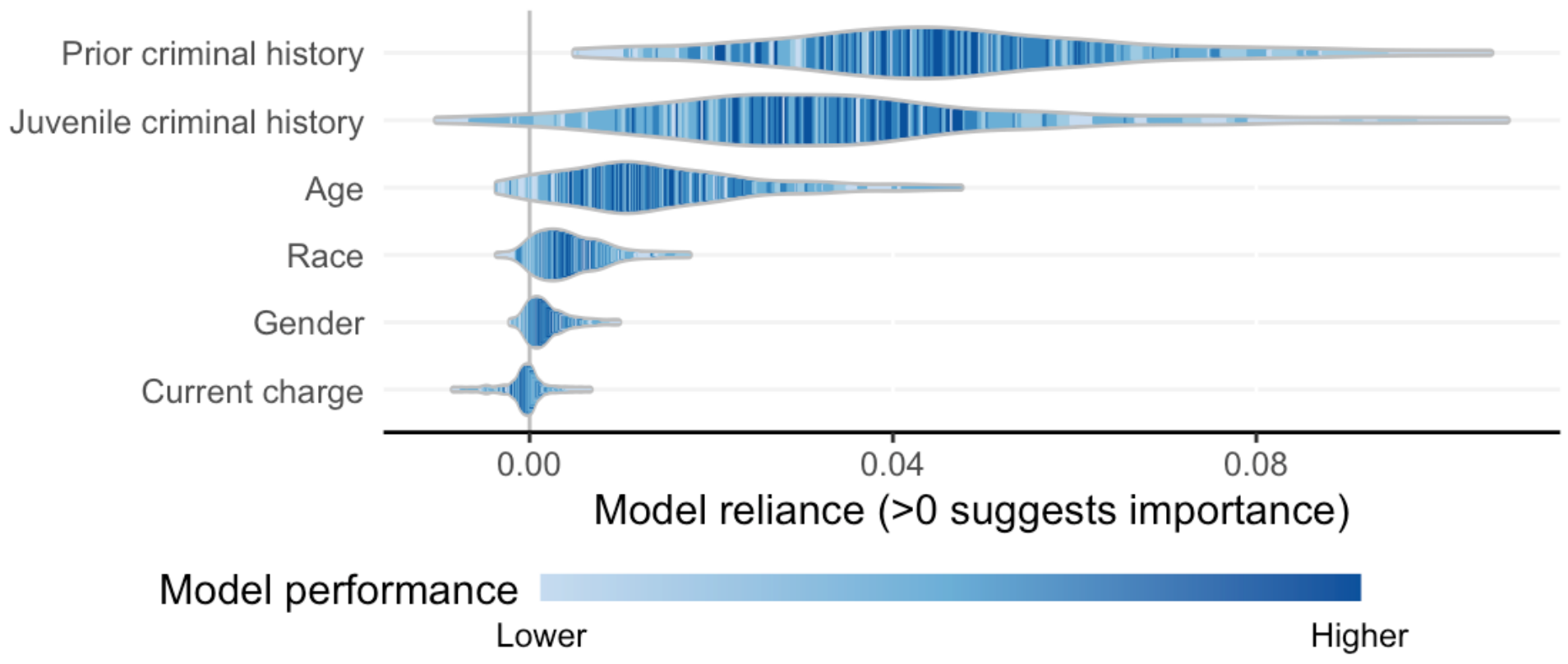

**Figure A2.** Spearman correlation among variables in the MIMIC study, arranged by variance inflation factor from the minimum-loss model. Absolute values larger than 0.4 are colored (in blue for positive correlation and red for negative correlation).

| | Glucose | WBC | SpO2 | Platelet | Temperature | Respiration rate | Age | Potassium | Lactate | Heart rate | BUN | Creatinine | SBP | DBP | Anion gap | MAP | Sodium | Bicarbonate | Chloride | Hemoglobin |
|---|---|---|---|---|---|---|---|---|---|---|---|---|---|---|---|---|---|---|---|---|
| WBC | 0.14 | | | | | | | | | | | | | | | | | | | |
| SpO2 | 0.01 | 0.03 | | | | | | | | | | | | | | | | | | |
| Platelet | 0.04 | 0.29 | -0.05 | | | | | | | | | | | | | | | | | |
| Temperature | 0.01 | 0.12 | 0.06 | 0 | | | | | | | | | | | | | | | | |
| Respiration rate | 0.05 | 0.11 | -0.25 | 0.09 | 0.12 | | | | | | | | | | | | | | | |
| Age | 0.07 | 0.01 | -0.06 | -0.05 | -0.16 | 0.07 | | | | | | | | | | | | | | |
| Potassium | 0.08 | 0.08 | 0.01 | -0.03 | -0.07 | -0.05 | 0.08 | | | | | | | | | | | | | |
| Lactate | 0.12 | 0.12 | 0.08 | -0.11 | 0.02 | 0.01 | -0.02 | 0.05 | | | | | | | | | | | | |
| Heart rate | 0.08 | 0.17 | -0.05 | 0.03 | 0.27 | 0.29 | -0.21 | 0 | 0.1 | | | | | | | | | | | |
| BUN | 0.14 | 0.03 | -0.07 | -0.08 | -0.18 | 0.11 | 0.34 | 0.3 | 0.03 | -0.07 | | | | | | | | | | |
| Creatinine | 0.1 | 0.01 | -0.06 | -0.08 | -0.11 | 0.08 | 0.17 | 0.29 | 0.07 | -0.08 | 0.74 | | | | | | | | | |
| SBP | 0.08 | -0.06 | 0.01 | 0.07 | 0.05 | -0.03 | 0 | -0.08 | -0.04 | -0.09 | -0.03 | 0 | | | | | | | | |
| DBP | 0.01 | -0.07 | -0.03 | 0.07 | 0.02 | 0.01 | -0.36 | -0.11 | 0 | 0.2 | -0.2 | -0.11 | 0.49 | | | | | | | |
| Anion gap | 0.16 | 0.11 | -0.07 | 0.14 | -0.04 | 0.16 | -0.02 | 0.1 | 0.18 | 0.09 | 0.34 | 0.42 | 0.07 | 0.07 | | | | | | |
| MAP | 0.06 | -0.04 | 0.03 | 0.06 | 0.05 | -0.03 | -0.22 | -0.11 | 0.01 | 0.09 | -0.16 | -0.1 | 0.76 | 0.85 | 0.07 | | | | | |
| Sodium | -0.06 | -0.04 | 0.01 | 0 | 0.03 | 0.02 | 0 | -0.19 | 0.01 | -0.05 | 0.01 | 0 | 0.11 | 0.08 | 0 | 0.1 | | | | |
| Bicarbonate | -0.06 | -0.13 | -0.12 | 0.09 | -0.01 | -0.07 | 0.04 | -0.11 | -0.17 | -0.13 | -0.13 | -0.18 | 0.11 | 0.05 | -0.42 | 0.06 | 0.11 | | | |
| Chloride | -0.08 | 0.03 | 0.19 | -0.19 | 0.06 | -0.07 | 0 | -0.01 | 0.05 | 0.01 | -0.09 | -0.1 | -0.07 | -0.06 | -0.24 | -0.04 | 0.54 | -0.41 | | |
| Hemoglobin | 0.02 | 0.08 | -0.13 | 0.09 | 0.03 | -0.02 | -0.17 | -0.08 | 0.06 | -0.07 | -0.23 | -0.11 | 0.13 | 0.29 | 0.12 | 0.25 | 0.1 | 0.08 | -0.1 | |
| Hematocrit | 0.03 | 0.1 | -0.15 | 0.13 | 0.01 | 0 | -0.15 | -0.04 | 0.06 | -0.06 | -0.2 | -0.08 | 0.12 | 0.28 | 0.14 | 0.23 | 0.12 | 0.1 | -0.11 | 0.97 |

BUN: blood urea nitrogen; DBP: diastolic blood pressure; MAP: mean arterial pressure; SBP: systolic blood pressure; $SpO_2$: peripheral capillary oxygen saturation; WBC: white blood cell.

**Figure A3.** Visual summary of MIMIC study results from the VIC analysis. VIC values were subtracted 1 such that zero value or lower indicate unimportance. (A) Average values from the 350 nearly optimal models. Five of the top seven variables (i.e., except MAP and chloride) by average VIC values were also ranked top seven by average ShapleyVIC values. (B) Distribution of VIC values (subtracted 1) and the corresponding model performance.

**(A)** Variable ranking based on average VIC values (subtracted 1) from 350 nearly optimal models.

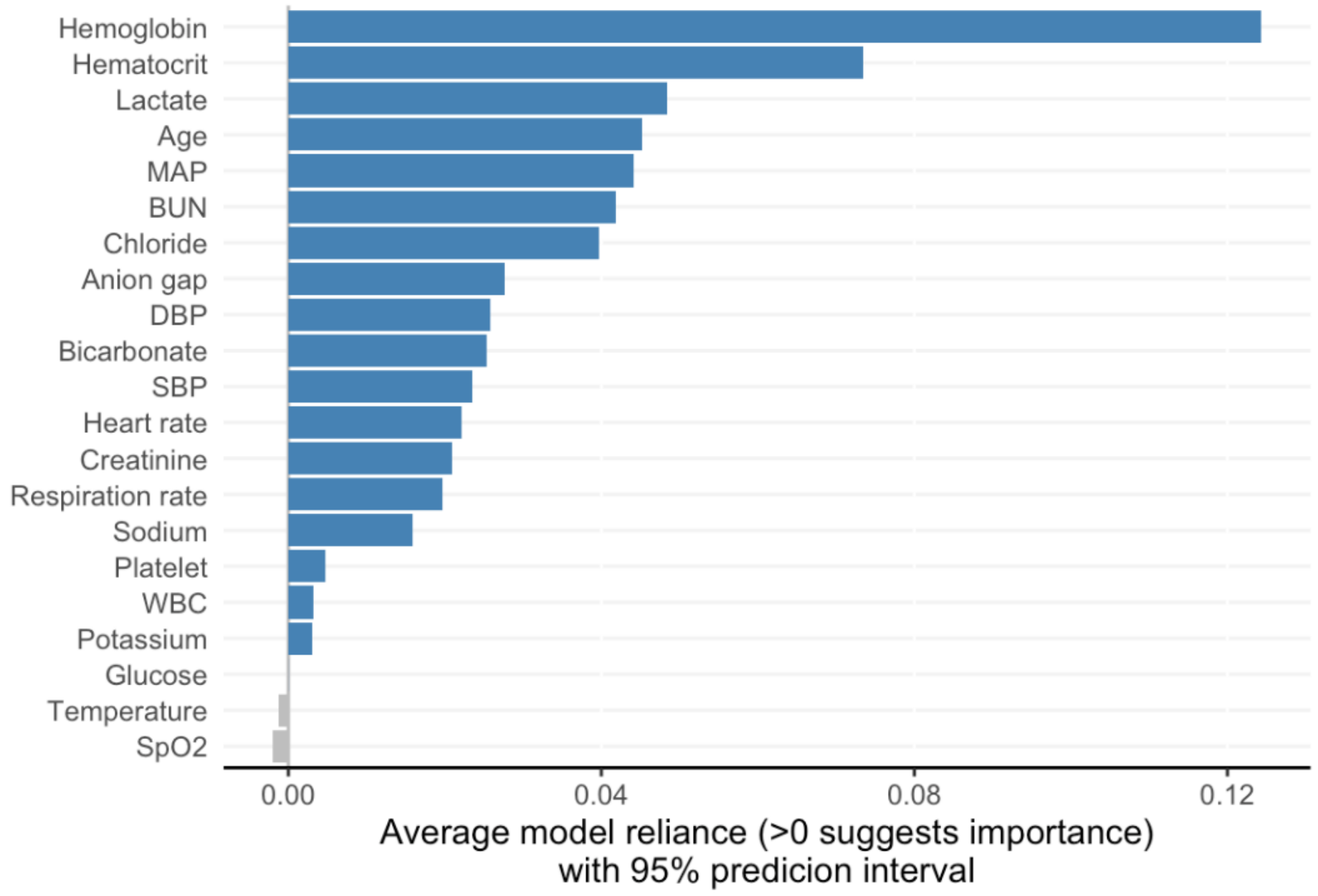

**(B)** VIC values (subtracted 1) from 350 nearly optimal models.

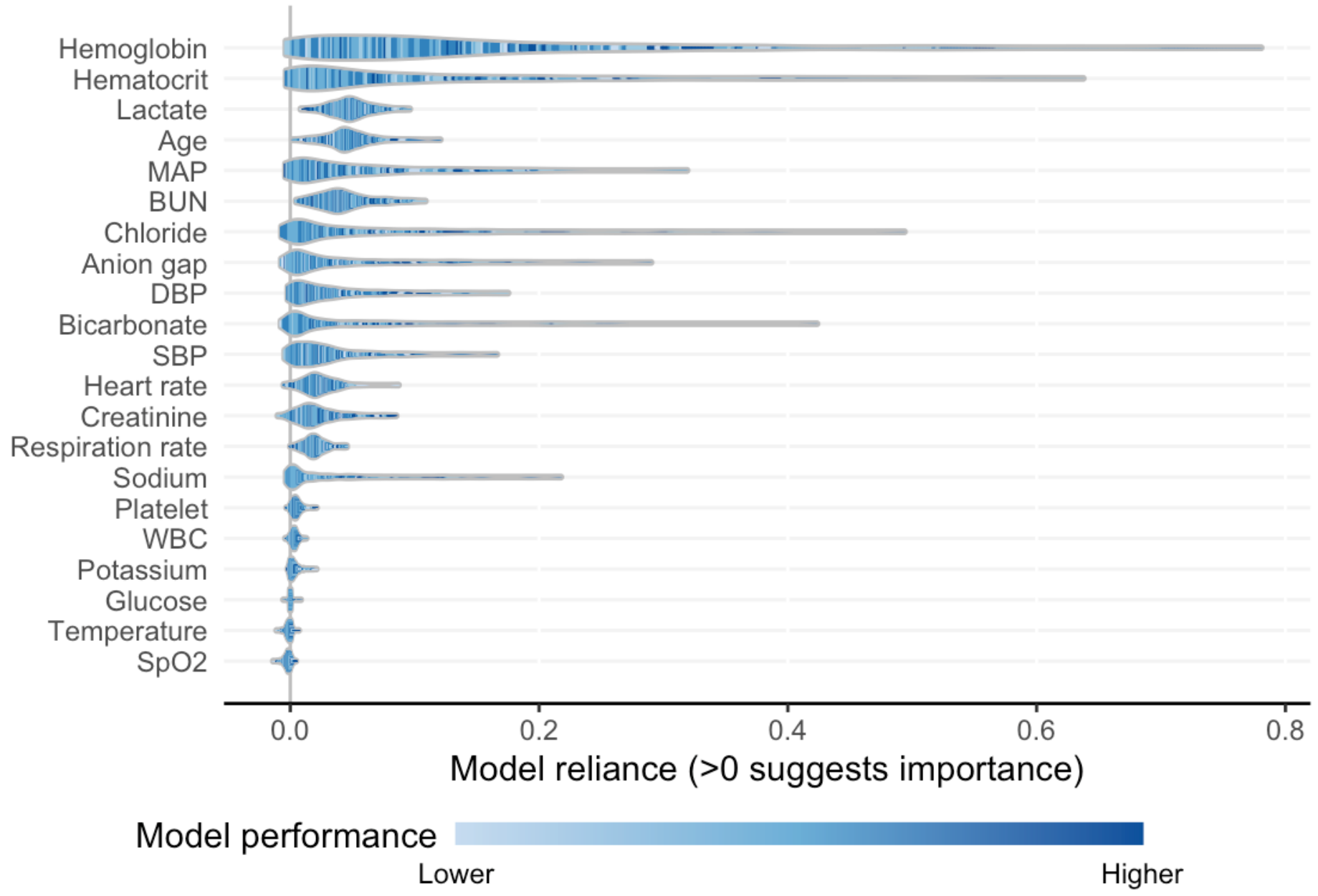

BUN: blood urea nitrogen; DBP: diastolic blood pressure; MAP: mean arterial pressure; SBP: systolic blood pressure; $SpO_2$: peripheral capillary oxygen saturation; WBC: white blood cell.